\documentclass[11pt]{article}

\usepackage[preprint]{acl}

\usepackage{times}
\usepackage{latexsym}

\usepackage[T1]{fontenc}

\usepackage[utf8]{inputenc}

\usepackage{microtype}

\usepackage{inconsolata}

\usepackage{graphicx}

\usepackage[table]{xcolor}
\usepackage{booktabs}
\usepackage{fontawesome5}
\usepackage{multirow}

\usepackage{array}
\usepackage{caption}
\usepackage{geometry}
\usepackage{pifont} 

\title{MMLongEmbed: Benchmarking Multimodal Embedding Models in Long-Context Scenarios}

\author{
  Haitian Wang\textsuperscript{\rm 1}, 
  Ruoxi Sun\textsuperscript{\rm 1}, 
  Quantong Qiu\textsuperscript{\rm 1}, 
  Juntao Li\textsuperscript{\rm 1}\thanks{Corresponding author: \texttt{ljt@suda.edu.cn}}, 
  Junhui Li\textsuperscript{\rm 1}, \\
  \textbf{Hua Chen\textsuperscript{\rm 2}, 
  Jinxiong Chang\textsuperscript{\rm 2}, 
  Min Zhang\textsuperscript{\rm 1}} \\
  \\
  \textsuperscript{\rm 1}Soochow University, Suzhou, China \\
  \textsuperscript{\rm 2}Ant Group, Beijing, China \\
}
\begin{document}
\maketitle

\begin{abstract}
Recent advancements have significantly expanded the theoretical context windows of Multimodal Embedding Models (MEMs). However, larger context windows do not necessarily translate into effective comprehension and representation of long-context multimodal inputs, which remains a critical bottleneck for real-world deployment. To address the lack of systematic evaluation in this setting, we introduce \emph{MMLongEmbed}, the first comprehensive benchmark for evaluating MEMs in long-context scenarios. MMLongEmbed comprises four retrieval tasks spanning multiple context-length ranges, covering text, document, and video modalities. Through extensive evaluation of state-of-the-art models, we find that current architectures rely heavily on superficial feature matching and struggle to capture deep semantic and structural dependencies. We further observe that performance degradation varies systematically with context length and key information placement. Moreover, models exhibit substantially different robustness to redundant contextual information across modalities. For reproducibility, the benchmark and code are publicly available.\footnote{\url{https://github.com/AmamiSora1228/MMLongEmbed}}
\end{abstract}

\section{Introduction}
In recent years, the context windows of Multimodal Embedding Models (MEMs) have significantly expanded to 32K tokens and beyond, enabling unified representations of long multimodal contexts interleaving text, document images, and videos~\cite{li2026qwen3,meng2025vlm2vecv2}. This capability has driven the development of various downstream applications, including multimodal retrieval-augmented generation (RAG)~\cite{faysse2025colpali,meng2025multimodal,zhang2026reasoning}, autonomous agent workflows~\cite{gulati2026beyond}, and large-scale cross-modal semantic search. Meanwhile, recent work has increasingly moved beyond superficial feature matching, incorporating implicit reasoning processes during representation learning to improve deep semantic embeddings~\cite{jiang2026embed,wang2026mmemb}.
\begin{figure}[t]
  \centering
    \includegraphics[width=\linewidth]{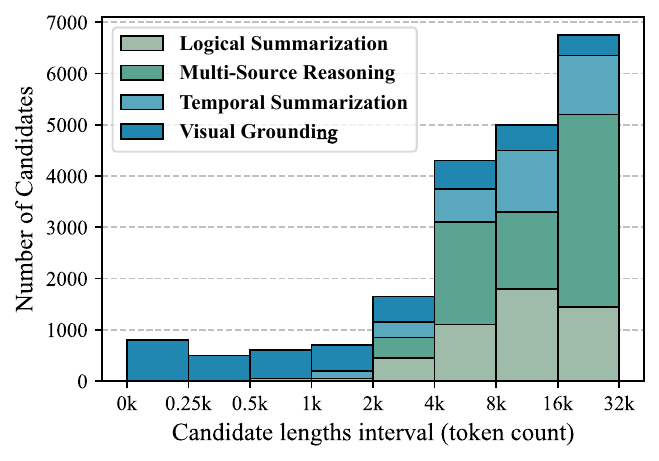}
    \caption{Length Distribution of Candidates in the MMLongEmbed Benchmark.}
  \label{fig:length_distribution}
\end{figure}

However, in stark contrast to the rapid evolution of model capabilities, evaluation benchmarks for MEMs significantly lag behind. Existing embedding benchmarks predominantly focus on short inputs and simple understanding tasks~\cite{xiao2025mieb,jiang2024vlm2vec,enevoldsen2025mmteb}. Meanwhile, the few existing embedding benchmarks oriented toward long inputs~\cite{zhu2024longembed} still suffer from several limitations: (1) \textbf{Limited Context Diversity:} they are restricted to text-only data, failing to cover various types of multimodal information encountered in real-world scenarios; (2) \textbf{Lack of Length Stratification Design:} input length is not treated as a controlled variable, preventing systematic analysis of performance variation as context scales; and (3) \textbf{Insufficient Task Difficulty:} task difficulty is not explicitly controlled, with candidate pools simply inflated using all available data, resulting in overly easy benchmarks where many tasks can be solved via shallow feature matching.

To tackle these deficiencies, we introduce \emph{MMLongEmbed}, a comprehensive benchmark for evaluating multimodal embedding models in long-context scenarios, bridging key gaps in existing evaluation frameworks. Specifically, to overcome the first limitation, MMLongEmbed incorporates four distinct data sources: documents, webpages, video frame sequences, and long-form text, and supports three context formats: image-only, interleaved image--text, and text-only, as summarized in Table~\ref{tab:dataset_stats}. To address the second limitation, we explicitly treat input length as a controlled variable through a length-stratified pooling mechanism, where candidate lengths extend up to 32K tokens. To address the third deficiency, we increase retrieval difficulty beyond simple candidate pool expansion by constructing high-quality distractors. For synthetic tasks, we select highly similar needle distractors to introduce fine-grained retrieval ambiguity. For real-world tasks, we introduce semantically challenging distractors that require models to transcend superficial feature matching.

We benchmark 11 leading MEMs, including 10 open-source models and one proprietary model, to provide a comprehensive evaluation. Experimental results show that current MEMs still face substantial challenges in long-context scenarios, and that increasing model size or embedding dimensionality does not consistently improve performance. We further observe a clear degradation trend in fine-grained retrieval performance as context length increases, while performance on summarization-style tasks remains relatively stable, likely due to higher redundancy and more distributed evidence within the context. Furthermore, when introducing targeted distractors to suppress superficial matching cues, we find that existing MEMs heavily rely on surface-level semantic alignment and exhibit clear limitations in tasks requiring deeper understanding.

\section{Related Work}
\label{sec:related_work}

\begin{table*}[t]
    \centering
    \resizebox{\textwidth}{!}{
        \begin{tabular}{@{} l c c c c ccc @{}}
            \toprule
            \multirow{2.5}{*}{\textbf{Dataset}} & \multicolumn{4}{c}{\textbf{Dataset Information}} & \multicolumn{3}{c}{\textbf{Statistics of Datasets}} \\
            \cmidrule(lr){2-5} \cmidrule(lr){6-8}
            & \textbf{Source} & \textbf{Corpus Type} & \textbf{Corpus Modality} & \textbf{Task type} & \textbf{Query} & \textbf{Corpus} & \textbf{Avg. C/Q} \\
            \midrule

            \rowcolor{gray!15} \multicolumn{8}{@{}c}{\textbf{\textit{Visual Grounding}}} \\
            NeedleBench         & \citep{li2024needlebenchllmsretrievalreasoning}                  & WebPage  & Image+Text & Synthetic Tasks & 2160 & 2160 & 25 \\
            MM-NIAH              & \citep{wang2024needle}                                           & WebPage  & Image+Text & Synthetic Tasks & 2160 & 2160 & 25 \\
            \midrule

            \rowcolor{gray!15} \multicolumn{8}{@{}c}{\textbf{\textit{Multi-Source Reasoning}}} \\
            MMDocRAG            & \citep{dong2025benchmarking}                                     & Document & Image+Text & Real Tasks      & 960  & 3840 & 640 \\
            HotpotQA            & \citep{yang2018hotpotqa}                                  & Text     & Text       & Real Tasks      & 960  & 3840 & 640 \\
            \midrule

            \rowcolor{gray!15} \multicolumn{8}{@{}c}{\textbf{\textit{Logical Summarization}}} \\
            MMLongBench         & \citep{wang2025mmlongbenchbenchmarkinglongcontextvisionlanguage} & Document & Image      & Real Tasks      & 480  & 1920 & 320 \\
            GovReport           & \citep{Huang2021EfficientAF}                                         & Text     & Text       & Real Tasks      & 720 & 2880 & 480 \\
            \midrule

            \rowcolor{gray!15} \multicolumn{8}{@{}c}{\textbf{\textit{Temporal Summarization}}} \\
            ActivityNet Captions & \citep{krishna2017dense}                                      & Video    & Image      & Real Tasks      & 420  & 1680 & 280 \\
            YoukuDenseCaption             & \citep{xiong2025youku}                                           & Video    & Image      & Real Tasks      & 600  & 2400 & 400 \\

            \bottomrule
        \end{tabular}
    }
    \caption{Statistics of Tasks in MMLongEmbed.}
    \label{tab:dataset_stats}
\end{table*}
\paragraph{Multimodal Embedding Models}
MEMs aim to map heterogeneous modalities into a unified embedding space for cross-modal alignment. Early work primarily relied on contrastive dual-encoder architectures, which establish a shared semantic space between vision and language through large-scale image--text pretraining~\cite{clip,align}. Building on this foundation, recent advances driven by multimodal large language models (MLLMs) have increasingly focused on learning more general and flexible embedding spaces, enabling richer multimodal representations and broader application scenarios~\cite{jiang2024e5vuniversalembeddingsmultimodal,meng2025vlm2vecv2,zhang2024gme,lin2025mm}. More recent developments further improve representation quality and robustness in more challenging settings, including stronger query--target alignment and more efficient multimodal encoding~\cite{li2026magicmmembeddingvisualtokenefficientuniversalmultimodal,wang2026mmemb}. In addition, some MEMs have begun to support larger input windows, enabling document-level multimodal understanding~\cite{li2026qwen3}. These advances have made MEMs a key component in emerging applications such as multimodal RAG, visual document retrieval, and autonomous agent workflows~\cite{faysse2025colpali}, which inherently require processing extensive contexts. However, despite these substantial improvements, the systematic evaluation of MEMs in long-context settings remains limited.

\paragraph{Benchmarking Embedding Models}
Benchmarking of embedding models has evolved from short-text retrieval to a comprehensive evaluation framework covering multiple domains, languages, task types, and modalities. Early benchmarks established a unified zero-shot evaluation paradigm for cross-domain text retrieval~\cite{thakur2021beir}, later extended to multilingual and cross-lingual retrieval tasks~\cite{zhang2022making,winata2024miners}. Subsequent benchmarks further expanded to comprehensive assessments of embedding capabilities, covering multi-task text embedding, multilingual evaluation, and long-horizon memory retrieval~\cite{muennighoff2023mteb,enevoldsen2025mmteb,zhao2026lmeb}. Meanwhile, embedding benchmarks have extended to long-input settings, laying the foundation for systematic assessment of representation capabilities in long-context tasks~\cite{zhu2024longembed}. In parallel, a growing number of benchmarks have introduced image--text and broader multimodal embedding tasks, expanding cross-modal coverage~\cite{jiang2024vlm2vec,xiao2025mieb}. However, current evaluation frameworks remain incomplete in two key aspects: existing multimodal embedding benchmarks are largely limited to short contexts, and long-context embedding evaluations focus primarily on text-only scenarios, leaving multimodal embedding under long-context conditions largely unexplored. To fill this gap, we introduce \emph{MMLongEmbed}, the first comprehensive benchmark designed to systematically evaluate MEMs in long-context scenarios. A detailed comparison with existing embedding benchmarks is provided in Table~\ref{tab:benchmark_comparison}.

\section{MMLongEmbed}
\label{sec:benchmark}

\begin{figure*}[t]
    \centering
    \includegraphics[width=\textwidth]{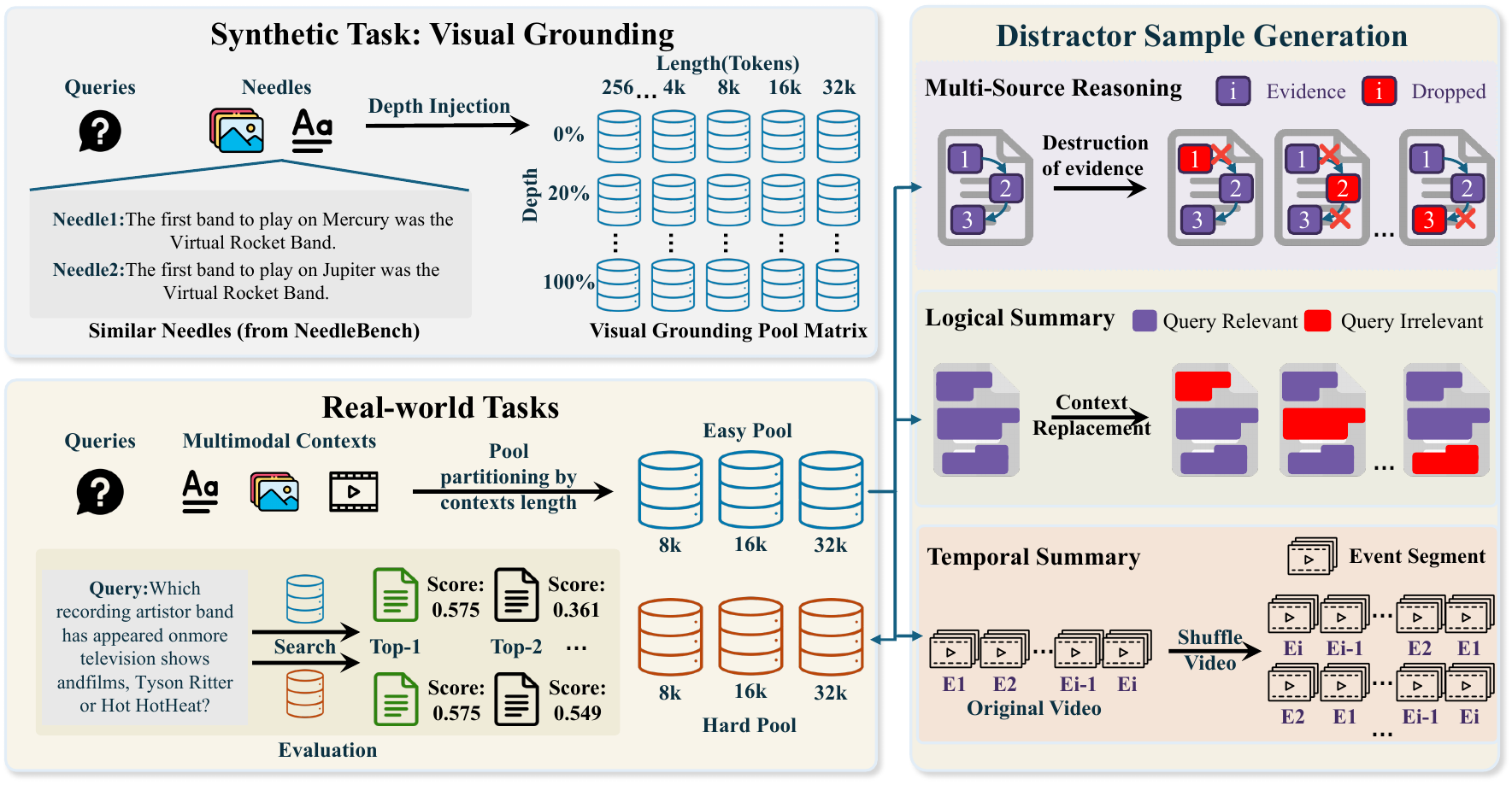}
    \caption{\textbf{Overview of the MMLongEmbed data construction pipeline.} For synthetic visual grounding, we inject target instances (``needles'') across varying insertion depths and expanding context lengths, with each depth--length configuration evaluated independently. For real-world scenarios, we construct parallel hard pools utilizing targeted distractors to effectively prevent models from exploiting superficial semantic matching shortcuts. Furthermore, these real-world tasks are strictly stratified into discrete length intervals (8K, 16K, and 32K). \textit{Note:} ``Temporal Summary'' and ``Logical Summary'' are abbreviated labels in the figure due to space constraints; the full task names are Temporal Summarization and Logical Summarization, respectively.}
    \label{fig:pipeline}
\end{figure*}
Our benchmark, MMLongEmbed, comprises four distinct tasks and supports three input formats: image-only, interleaved image--text, and text-only. It integrates eight diverse datasets, with a total of 8,460 queries and 20,880 candidates, and supports context lengths of up to 32K tokens. Detailed dataset statistics are provided in Table~\ref{tab:dataset_stats}.

\subsection{Query and Corpus Setting}
To systematically investigate the boundaries of model capabilities under increasing input length, we introduce a strictly length-stratified pooling mechanism. In our setup, all queries are relatively short, while the corresponding multimodal candidates are significantly longer. Accordingly, we partition these query-candidate pairs into separate pools based on the candidate length. Specifically, synthetic tasks span eight length intervals, whereas real-world tasks are divided into three (8K, 16K, and 32K). Each length interval is evaluated independently. Furthermore, we maintain strict control over the candidate density; for a given task, the average number of candidates per query (Avg. C/Q) is kept identical across all length pools, while differing only across distinct tasks, with detailed statistics presented in Table~\ref{tab:dataset_stats}.

Since visual tokenization varies substantially across models, we utilize a token counting scheme as a standardized baseline to define consistent length intervals. Building upon MMLongBench~\cite{wang2025mmlongbenchbenchmarkinglongcontextvisionlanguage}, we revise the unified multimodal token counting scheme to align with recent MEMs~\cite{wang2025internvl3,tschannen2025siglip}. The formulation is as follows:

\begin{itemize}
    \item \textbf{Text tokens}: Computed using the Qwen3 tokenizer~\cite{bai2025qwen3}.

    \item \textbf{Visual tokens}: We adopt a $16 \times 16$ patch embedding followed by a $2 \times 2$ spatial merging operation, yielding an effective patch size of $32 \times 32$ pixels per visual token. Accordingly, for an image of resolution $H \times W$, the number of visual tokens is estimated as:
    \begin{equation}
        N_{\mathrm{image}} = \left\lceil \frac{H}{32} \right\rceil \times \left\lceil \frac{W}{32} \right\rceil.
    \end{equation}
\end{itemize}

The total context length $L$ is defined as the total number of tokens after multimodal tokenization:
\begin{equation}
    L = N_{\mathrm{text}} + \sum_{i=1}^{K} N_{\mathrm{image}}^{(i)},
\end{equation}
where $N_{\mathrm{text}}$ denotes the total number of text tokens aggregated over all text segments in the sequence, $N_{\mathrm{image}}^{(i)}$ denotes the visual token count of the $i$-th image, and $K$ is the number of images contained in the sequence.

\subsection{Task Setup and Benchmark Construction}
MMLongEmbed comprises four evaluation tasks constructed from eight public datasets originally designed for generative evaluation, categorized into one synthetic task (Visual Grounding) and three real-world tasks (Multi-Source Reasoning, Logical Summarization, and Temporal Summarization). These tasks redefine the source datasets as retrieval problems, evaluating the ability of MEMs to deeply understand and represent long-context multimodal information. We defer the detailed construction procedure to Figure~\ref{fig:pipeline} and present the design of each task in the following sections. Task details are in Appendix~\ref{app:task_descriptions}.

\subsubsection{Visual Grounding}

To systematically evaluate the fine-grained information retention capabilities of MEMs, we construct a multimodal ``Needle-In-A-Haystack'' evaluation framework. Specifically, we curate 25 query--needle pairs for both text and image modalities from NeedleBench~\cite{li2024needlebenchllmsretrievalreasoning} and MM-NIAH~\cite{wang2024needle}, respectively. These needles are embedded into multimodal haystacks sampled from the OBELICS dataset~\cite{laurenccon2023obelics}. Context lengths range from 256 to 32,768 tokens across eight intervals, while insertion depths are uniformly discretized into six levels spanning 0\% to 100\%.

For each length--depth configuration, we construct a dedicated corpus comprising 25 candidate documents. Each document contains a needle instance, and all needle instances are designed to exhibit high semantic and structural similarity to each other. This setup forms a controlled retrieval environment characterized by strong instance-level confusability across candidates.

\subsubsection{Multi-Source Reasoning}

This task assesses the models' capability to capture structured evidence required for multi-source reasoning within heterogeneous contexts. Built upon HotpotQA~\cite{yang2018hotpotqa} and MMDocRAG~\cite{dong2025benchmarking}, we filter out overly generic queries and select instances containing 2--5 evidence segments, with original contexts up to 8K tokens. We then expand these contexts to 16K and 32K tokens by injecting semantically irrelevant yet naturally flowing textual context, while preserving the relative positions of key evidence segments.

Hard interference samples are generated by corrupting these expanded candidates, removing critical evidence spans (e.g., the first, last, or random intermediate segments). This yields variants that share surface-level topical similarity with the query but lack complete reasoning chains. A strictly size-matched easy pool is additionally constructed by padding other documents to uniform lengths.

\subsubsection{Logical Summarization}

This task evaluates global semantic compression by demanding precise alignment between concise target summaries and long-range narrative documents. We utilize GovReport~\cite{Huang2021EfficientAF} and MMLongBench~\cite{wang2025mmlongbenchbenchmarkinglongcontextvisionlanguage}, filtering for subsets that require macro-level document understanding rather than simple span extraction. 

We construct challenging variants by replacing large contiguous spans (e.g., the first 25\%, last 25\%, or middle 50\% of the text) with query-irrelevant content. Crucially, the injected text is sampled from topically similar documents within the same corpus, ensuring high lexical overlap and surface fluency. This mechanism fundamentally breaks global semantic alignment while preserving local grammatical coherence, explicitly forcing models to look beyond localized matching cues. For the easy pool, unmodified documents of equivalent lengths are used.

\subsubsection{Temporal Summarization}

Parallel to the logical summarization task, this task evaluates the ability to summarize and understand long-range temporal structure in video data, sourced from YoukuDenseCaption (YoukuDC)~\cite{xiong2025youku} and ActivityNet Captions~\cite{krishna2017dense}. During the data curation pipeline, all raw videos are uniformly downsampled to 1 FPS to construct standardized temporal sequences.

For this task, hard distractors are generated via event-level temporal perturbations. Instead of naive frame shuffling, videos are segmented at semantic boundaries and subjected to macro-structural transformations, including chronological reversal and block-level permutation. These operations preserve event-level semantic units while disrupting global narrative structure and temporal ordering. This design exposes models that rely on order-agnostic or shallow visual similarities, thereby probing their ability to capture long-range temporal structure. For the easy pool, other unmodified videos are used without temporal perturbations.

\begin{table}[t]
    \centering
    \resizebox{\linewidth}{!}{%
    \begin{tabular}{lccc}
        \toprule
        \textbf{Model} & \textbf{Length} & \textbf{Size} & \textbf{Dim} \\
        \midrule
        
        \rowcolor{gray!20} \multicolumn{4}{@{}l@{}}{\textit{Open-Source / Local Models}} \\
        Qwen3-VL-2B~\citep{li2026qwen3}       & 32768 & 2B & 2048 \\
        Qwen3-VL-8B~\citep{li2026qwen3}       & 32768 & 8B & 4096 \\
        GME-Qwen2-VL-2B~\citep{zhang2024gme}  & 32768 & 2B & 1536 \\
        GME-Qwen2-VL-7B~\citep{zhang2024gme}  & 32768 & 7B & 3584 \\
        Rzen-v2-7B~\citep{jian2025rzenembed}  & 32768 & 7B & 3584 \\
        Ops-MM-v1-2B~\cite{ops_mm_embedding_2b} & 32768 & 2B & 2048 \\ 
        Ops-MM-v1-7B~\cite{ops_mm_embedding_7b} & 32768 & 7B & 3584 \\
        Embed-RL-2B~\citep{jiang2026embed}    & 32768 & 2B & 1536 \\
        Embed-RL-4B~\citep{jiang2026embed}    & 32768 & 4B & 2560 \\
        VLM2Vec-Qwen2VL-2B~\citep{jiang2024vlm2vec} & 32768 & 2B & 1536 \\
        
        \midrule
        \rowcolor{gray!20} \multicolumn{4}{@{}l@{}}{\textit{Proprietary Models}} \\
        Doubao-Seed-1.6~\cite{seed2025seed1}  & 131072 & {\scriptsize \faLock} & 2048 \\
        
        \bottomrule
    \end{tabular}%
    }
    \caption{Detailed model configurations. \faLock~indicates undisclosed proprietary model sizes.}
    \label{tab:model_details}
\end{table}
\begin{table*}[t]
    \centering
    \resizebox{\textwidth}{!}{
        \begin{tabular}{@{} l c c c >{\columncolor{gray!15}}c c >{\columncolor{gray!15}}c c >{\columncolor{gray!15}}c c >{\columncolor{gray!15}}c c >{\columncolor{gray!15}}c c >{\columncolor{gray!15}}c c @{}}
            \toprule
            \multirow{3.5}{*}{\textbf{Model}} & 
            \multicolumn{2}{c}{\textbf{Visual Grounding}} & 
            \multicolumn{4}{c}{\textbf{Multi-Source Reasoning}} & 
            \multicolumn{4}{c}{\textbf{Logical Summarization}} & 
            \multicolumn{4}{c}{\textbf{Temporal Summarization}} & 
            \multirow{3.5}{*}{\textbf{Avg}} \\
            \cmidrule(lr){2-3} \cmidrule(lr){4-7} \cmidrule(lr){8-11} \cmidrule(lr){12-15}
            
            & \textbf{NeedleBench} & \textbf{MM-NIAH} 
            & \multicolumn{2}{c}{\textbf{MMDocRAG}} & \multicolumn{2}{c}{\textbf{HotpotQA}} 
            & \multicolumn{2}{c}{\textbf{GovReport}} & \multicolumn{2}{c}{\textbf{MMLB}} 
            & \multicolumn{2}{c}{\textbf{YoukuDC}} & \multicolumn{2}{c}{\textbf{ActNet}} 
            & \\
            \cmidrule(lr){4-5} \cmidrule(lr){6-7} \cmidrule(lr){8-9} \cmidrule(lr){10-11} \cmidrule(lr){12-13} \cmidrule(lr){14-15}
            
            & - & - 
            & Easy & \multicolumn{1}{c}{Hard} & Easy & \multicolumn{1}{c}{Hard} 
            & Easy & \multicolumn{1}{c}{Hard} & Easy & \multicolumn{1}{c}{Hard} 
            & Easy & \multicolumn{1}{c}{Hard} & Easy & \multicolumn{1}{c}{Hard} 
            & \\
            \midrule
            
            Qwen3-VL-2B             & 48.70 & 41.62 & 43.75 & 32.29 & 61.46 & 60.00 & 81.84 & 65.56 & 86.25 & 12.50 & 76.00 & 13.00 & 89.33 & 20.00 & 52.31 \\
            Qwen3-VL-8B             & 48.84 & 35.46 & 43.96 & 25.42 & 66.67 & \underline{63.75} & 80.83 & 66.11 & 90.83 & 17.50 & \underline{83.00} & 11.33 & \underline{93.33} & \underline{20.67} & 53.41 \\
            GME-Qwen2-VL-2B         & 38.29 & 33.96 & 19.79 & 14.79 & 1.67 & 0.63 & 18.89 & 7.78 & 70.00 & 16.67 & 38.33 & 7.33 & 49.33 & 6.00 & 23.10 \\
            GME-Qwen2-VL-7B         & 35.42 & 35.69 & 18.33 & 14.58 & 2.29 & 0.63 & 21.67 & 8.89 & 70.83 & 19.17 & 43.33 & 12.67 & 51.33 & 10.00 & 24.63 \\
            Rzen-v2-7B              & \textbf{55.97} & \underline{59.17} & \textbf{60.21} & \textbf{54.58} & 73.33 & \textbf{65.63} & \underline{83.06} & \underline{69.72} & 90.83 & 11.67 & \textbf{87.67} & 16.00 & 72.67 & 20.00 & \textbf{58.61} \\
            Ops-MM-v1-7B            & \underline{52.18} & \textbf{60.69} & 42.29 & \underline{39.17} & \textbf{79.17} & 58.75 & \textbf{84.44} & 49.44 & \textbf{92.92} & 16.25 & 81.67 & \underline{16.33} & 66.00 & 15.33 & \underline{53.90} \\
            Ops-MM-v1-2B            & 51.81 & 48.08 & 29.38 & 26.46 & \underline{76.46} & 48.13 & 81.67 & 60.28 & 81.25 & 8.33 & 70.33 & 14.33 & 64.67 & 13.33 & 48.18 \\
            Embed-RL-2B             & 51.67 & 30.19 & 45.21 & 32.71 & 57.92 & 19.17 & 65.28 & 31.39 & 76.67 & \textbf{28.75} & 44.33 & 6.00 & 52.67 & 8.00 & 39.28 \\
            Embed-RL-4B             & 45.42 & 40.28 & \underline{48.75} & 33.54 & 60.63 & 20.83 & 68.06 & 31.94 & 77.50 & \underline{25.83} & 49.00 & 8.67 & 50.00 & 11.33 & 40.84 \\
            VLM2Vec-Qwen2VL-2B      & 47.36 & 38.50 & 16.46 & 16.25 & 0.42 & 0.83 & 20.00 & 23.89 & 82.92 & 8.33 & 72.67 & 14.33 & 80.67 & 12.00 & 31.04 \\
            Doubao-Seed-1.6~{\scriptsize \faLock}         & 43.43 & 55.56 & 27.29 & 24.38 & 52.08 & 50.63 & 81.39 & \textbf{70.83} & \underline{92.50} & 12.50 & 79.67 & \textbf{18.00} & \textbf{96.00} & \textbf{25.33} & 52.11 \\
            
            \bottomrule
        \end{tabular}
    }

    \caption{Precision@1 performance comparison of evaluated MEMs on the MMLongEmbed benchmark. The `Hard' columns are shaded in gray to denote subsets containing high-similarity distractors. To accommodate page width constraints, datasets are abbreviated as follows: ActNet (ActivityNet Captions) and MMLB (MMLongBench). Precision@1 exhibits the largest relative between-model variation among the evaluated metrics; complete results for Recall@$k$, MRR@$k$, and nDCG@$k$ are provided in Table~\ref{tab:full_metrics}.}
    \label{tab:model_benchmark}
\end{table*}
\section{Evaluation}
\subsection{Experimental Setup}

We comprehensively evaluate 11 recent MEMs, each featuring a context window of at least 32K tokens. This evaluation suite comprises 10 open-source models and one proprietary model. The open-source models primarily range in scale from 2B to 8B parameters, with embedding dimensions spanning from 1536 to 4096. The selected models exhibit significant diversity, covering early modality-alignment models based on contrastive learning, universal multimodal embedding models, and recent reasoning-driven multimodal embedding models. Furthermore, the proprietary model extends the maximum context length to 128K tokens. Comprehensive specifications for all models are provided in Table~\ref{tab:model_details}. To prevent out-of-memory (OOM) errors for inputs exceeding the context limit due to varying visual tokenization schemes, we apply middle truncation~\cite{bai2024longbench}.

\subsection{Overall Observation}
We present the full evaluation results on MMLongEmbed in Table~\ref{tab:model_benchmark}.

\paragraph{Performance of MEMs}
MEM performance reveals a clear decoupling between model scale and multimodal embedding capability in long-context scenarios. Within most model families, increasing model scale yields only marginal improvements, suggesting that factors beyond scale, such as model architecture and training strategy, may play an important role. This is evident in the Qwen3 series, where the 8B model only slightly outperforms the 2B variant (53.41 vs. 52.31) and even underperforms it on the Hard subset of MMDocRAG (25.42 vs. 32.29). Similar limited gains are observed in Embed-RL, whereas Ops-MM is a notable exception with a larger improvement from 2B to 7B. Statistical comparisons of model-scale effects are provided in Table~\ref{tab:scale_tests}.

Moreover, the substantial performance gap between model families, such as Qwen3 and GME, suggests that scale alone does not determine long-context embedding performance. Overall, without targeted long-context optimization, increasing model capacity and embedding dimensionality appears insufficient for robust fine-grained discrimination in long-context scenarios.

\begin{table}[htbp]
\centering
\small
\renewcommand{\arraystretch}{1.2}
\setlength{\tabcolsep}{3pt}

\resizebox{\linewidth}{!}{
\begin{tabular}{
>{\raggedright\arraybackslash}p{0.36\linewidth}
>{\centering\arraybackslash}p{0.11\linewidth}
>{\centering\arraybackslash}p{0.11\linewidth}
>{\centering\arraybackslash}p{0.11\linewidth}
>{\centering\arraybackslash}p{0.16\linewidth}
>{\centering\arraybackslash}p{0.16\linewidth}
}
\toprule

\textbf{Model} 
& \textbf{Text} 
& \textbf{Mix} 
& \textbf{Image} 
& \textbf{Mean ($\mu$)} 
& \textbf{Std. Dev. ($\sigma$)} \\

\midrule

\rowcolor{gray!15}
\multicolumn{6}{@{}l@{}}{\textit{Category 1: Modality-Skewed Models}} \\
Doubao-Seed-1.6 & 75.83 & 87.15 & 85.14 & 82.71 & 6.03  \\
Qwen3-VL-2B     & 63.19 & 82.85 & 88.61 & 78.22 & 13.33 \\
Qwen3-VL-8B     & 64.17 & 81.67 & 89.03 & 78.29 & 12.77 \\
Ops-MM-v1-2B    & 75.28 & 80.14 & 88.91 & 81.44 & 6.91  \\
GME-Qwen2-VL-7B & 55.69 & 62.08 & 72.99 & 63.59 & 8.75  \\

\midrule

\rowcolor{gray!15}
\multicolumn{6}{@{}l@{}}{\textit{Category 2: Cross-Modal Interference Sensitive Models}} \\

Embed-RL-2B     & 65.49 & 61.94 & 73.61 & 67.01 & 6.00  \\
Embed-RL-4B     & 71.81 & 62.43 & 75.35 & 69.86 & 6.68  \\

\midrule

\rowcolor{gray!15}
\multicolumn{6}{@{}l@{}}{\textit{Category 3: Modality-Robust Models}} \\

Rzen-v2-7B      & 86.94 & 82.71 & 86.32 & 85.32 & 2.29  \\

\bottomrule
\end{tabular}
}

\caption{
Performance of MEMs under heterogeneous background interference in \emph{Visual Grounding}. The table reports Precision@1 under text-only, mixed-modal, and image-only distractors. Models are grouped by their performance patterns across the three distractor settings: \textit{modality-skewed} models show substantial variation across settings, \textit{interference-sensitive} models perform worse under mixed-modal than single-modal interference, and \textit{modality-robust} models remain relatively stable across settings. Mean ($\mu$) and sample standard deviation ($\sigma$) summarize overall performance and cross-setting variation, respectively.
}
\label{tab:modality_bias}

\end{table}
\paragraph{Performance Analysis on Easy Tasks}
The comparison across Easy task categories reveals clear differences in performance under long-context compression. \emph{Visual Grounding} is among the most challenging tasks for most models, and instance-level detail discrimination remains a major limitation for current MEMs in long-context tasks. \emph{Multi-Source Reasoning} exhibits moderate difficulty, requiring models to capture and integrate structured evidence across multiple sources. While overall performance remains constrained in multimodal settings, stronger results on the text-based HotpotQA dataset suggest that cross-modal alignment constitutes a major bottleneck,
whereas intrinsic reasoning capability appears less limiting in our setting. \emph{Summarization} tasks are comparatively easier, with many models exceeding 80\% precision even in video settings. A plausible explanation is that summarization-style queries admit global semantic matching, allowing models to aggregate contextual information without strict preservation of fine-grained evidence.

\begin{figure*}[t]
    \centering
    \includegraphics[width=\textwidth]{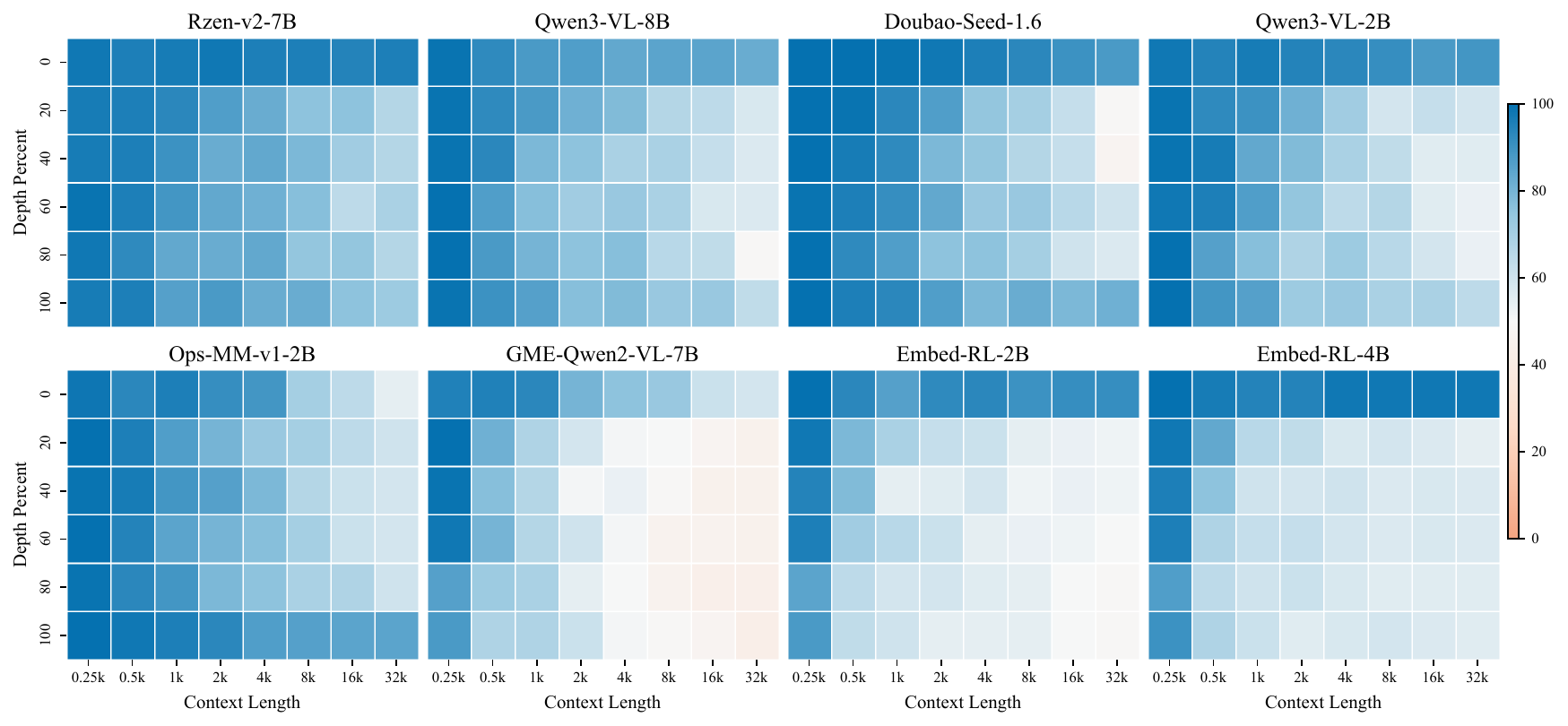}
    \caption{Precision@5 on the Visual Grounding task across varying insertion depths and context lengths.}
    \label{fig:heatmap_vg}
\end{figure*}

\section{Empirical Analysis}
\label{sec:empirical_analysis}

\begin{figure*}[htbp]
    \centering
    \includegraphics[width=\textwidth]{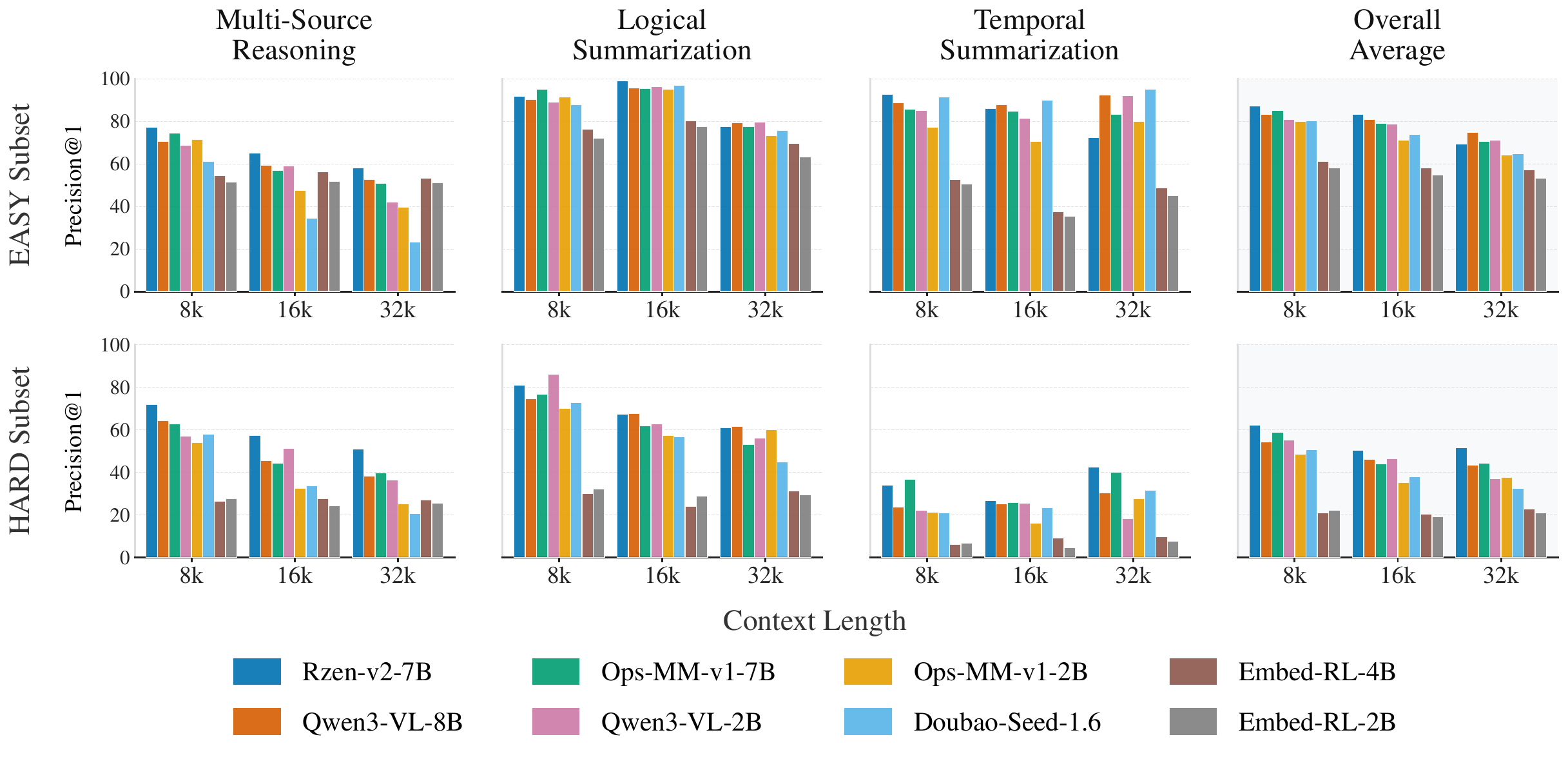}
    \caption{Quantitative comparison of model performance across context lengths (8K, 16K, 32K). Results are evaluated by Precision@1 on Multi-Source Reasoning, Logical Summarization, and Temporal Summarization under Easy and Hard settings, with the rightmost column reporting the overall average Precision@1.}
    \label{fig:length_with_task}
\end{figure*}

\paragraph{Performance Analysis on Hard Tasks}
A consistent performance gap emerges between Easy and Hard categories. While MEMs perform well in Easy conditions, their performance degrades sharply when superficial matching cues become ineffective due to high-similarity distractors. This finding suggests that models rely primarily on global semantic matching rather than deep fine-grained discrimination. The vulnerability is particularly pronounced in summarization tasks, where performance decreases substantially when distractors preserve partial semantic content while disrupting global structure. In contrast, Multi-Source Reasoning remains relatively stable under high-similarity interference, suggesting that dense embeddings can still capture coarse evidence chains to some extent, although they remain limited in assessing chain completeness and logical consistency. Temporal Summarization exhibits a distinct failure mode, with performance dropping sharply when temporal ordering is disrupted, indicating limited sensitivity to long-range event structure. Overall, these results show that MEMs can retrieve semantically relevant content under weak interference but struggle to maintain discriminative power when fine-grained structural distinctions become critical.

In this section, we systematically investigate the internal behavior and fundamental limitations of MEMs in long-context scenarios from three perspectives: (1) fine-grained information loss and positional bias (\S~\ref{sec:analysis_vg}); (2) robustness to heterogeneous cross-modal interference (\S~\ref{sec:analysis_modality}); and (3) performance dynamics across diverse real-world tasks as context length scales (\S~\ref{sec:analysis_realworld}).

To save space and avoid visual clutter, we focus on the eight representative MEMs, which represent a broad range of capabilities and architectural bottlenecks. Comprehensive results for all evaluated models are provided in Section~\ref{detailed_evaluation_results} of the supplementary material.

\subsection{Positional Bias and Fine-Grained Retention}
\label{sec:analysis_vg}
We compare MEM performance on the Visual Grounding task across varying context lengths and insertion depths in Figure~\ref{fig:heatmap_vg}. By varying these two factors independently, we can distinguish performance degradation associated with increasing context length from sensitivity to the position of target information. The resulting heatmaps reveal two key bottlenecks. First, we observe severe information loss in intermediate positions, accompanied by a strong positional bias toward sequence boundaries. When the target evidence is not located at the beginning or end of the sequence, retrieval performance drops substantially. This pattern reflects a clear non-uniformity in how target information at different sequence positions is represented and retrieved. This suggests that existing models struggle to preserve information uniformly across positions in long sequences. Statistical analyses of positional bias are provided in Table~\ref{tab:position_tests}. Second, the ability to retain fine-grained information degrades rapidly as context length increases. Most models maintain stable and precise retrieval performance only up to roughly 2K tokens, after which performance continuously deteriorates as sequences grow longer. This degradation is observed across multiple model families, indicating that the difficulty is not restricted to a particular architecture or model scale. Notably, models of different scales within the Qwen and Embed series exhibit highly similar heatmap patterns, further suggesting that increasing model scale or embedding dimensionality alone is insufficient to alleviate this issue.

\subsection{Robustness to Cross-Modal Interference}
\label{sec:analysis_modality}

We analyze how MEMs handle heterogeneous background interference in synthetic tasks by categorizing distractors into three types: text-only, image-only, and mixed-modal. As shown in Table~\ref{tab:modality_bias}, models exhibit different patterns across interference types. First, some models demonstrate strong modality skew: performance varies significantly across distractor types, indicating uneven robustness to modality-specific noise. Second, some models are highly sensitive to cross-modal interference. In these cases, compared with single-modality distractors, mixed-modal distractors lead to performance degradation, suggesting that these models have difficulty handling mixed inputs. Finally, a small subset of models exhibits modality-robust behavior, maintaining stable performance across all distractor types. This indicates improved alignment and better-balanced multimodal representation learning. Overall, these results suggest that modality robustness is highly non-uniform across models and that balanced multimodal alignment remains a key factor in robustness under heterogeneous interference.

\subsection{Performance Degradation under Length Scaling}
\label{sec:analysis_realworld}

Unlike synthetic tasks, these real-world tasks exhibit more heterogeneous performance dynamics as context length increases, as shown in Figure~\ref{fig:length_with_task}. Overall, performance in most real-world tasks does not undergo the abrupt collapse observed in synthetic tasks as context length increases, but instead follows a more gradual degradation pattern. This suggests that real-world tasks contain a higher density of salient evidence, making successful task completion less dependent on perfect fine-grained retention than in synthetic tasks.

\paragraph{Multi-Source Reasoning}
Performance generally declines as context length increases in both Easy and Hard settings. This steady yet persistent degradation becomes more pronounced under higher difficulty, indicating that while current models retain some multi-step retrieval and internal reasoning capability, their reliability deteriorates substantially as the context expands.

\paragraph{Logical Summarization}
The Easy setting exhibits a non-monotonic trend: performance improves at 16K before declining at 32K. This suggests that moderately longer contexts introduce more salient evidence that benefits summarization. Although excessively long contexts may contain even more relevant information, current models still struggle to fully process and integrate this evidence at 32K, leading to performance degradation. In the Hard setting, performance decreases steadily without a sharp collapse. This further suggests that, while current MEMs remain relatively strong on summarization tasks, they still require improvement in fine-grained discrimination.

\paragraph{Temporal Summarization}
This task presents the largest performance gap between the Easy and Hard settings. While performance remains relatively stable across context lengths in the Easy setting, performance in the Hard setting drops sharply for most models. This suggests that temporal discrimination is highly fragile under difficult distractors and that current MEMs still show clear limitations in reliably distinguishing temporal structure under increasingly complex long-context conditions.


\section{Conclusion}
We propose MMLongEmbed, a benchmark for systematically evaluating MEMs in long-context scenarios. Our results reveal a substantial gap between effective information utilization and representational capacity in long contexts. We find that current models rely heavily on surface-level feature matching and struggle to effectively capture deep semantic and long-range structural dependencies. Specifically, fine-grained information retention degrades significantly as context length increases. While performance remains relatively stable on summarization tasks, it drops sharply under distractor-based fine-grained discrimination settings. Overall, our findings indicate that scaling context length, model size, or embedding dimensionality alone is insufficient to address these limitations, underscoring the need for fundamentally improved long-context semantic modeling.
\section*{Limitations}
MMLongEmbed has several limitations. First, it currently evaluates context lengths up to 32K tokens, and the benchmark design is not directly extensible to arbitrarily longer inputs, which may limit evaluation as context windows continue to scale. Second, our benchmark focuses on text--image inputs and does not yet cover other modalities such as audio or continuous 3D spatial data. Finally, all datasets in MMLongEmbed are English-only, limiting the evaluation of multilingual long-context multimodal understanding.
\section*{Acknowledgments}
We thank the anonymous reviewers for their valuable comments. This work was supported by the National Natural Science Foundation of China (NSFC No. 62576232) and the Ant Group through the CCF-Ant Research Fund.
\bibliography{custom}
\clearpage
\appendix

\section{Task Descriptions}
\label{app:task_descriptions}

\subsection{Visual Grounding}
\label{subsec:visual_grounding}

To rigorously evaluate the fine-grained retrieval capabilities of multimodal embedding models, the Visual Grounding task adopts a ``Needle-In-A-Haystack'' (NIAH) paradigm with high-similarity distractors. This design prevents models from exploiting superficial matching heuristics and forces deep semantic and visual discrimination.

\paragraph{Textual Needles and Distractors}
For the textual modality, we construct contrastive needle pairs that share nearly identical syntactic structures and lexical overlap, differing only in core semantic entities. This ensures that the distractor acts as a strong localized source of interference. For example, given a specific retrieval query, the context bucket contains both the gold-standard target and a highly similar distractor:
\begin{itemize}
    \item \textbf{Target Needle:} \textit{``The first band to play on the Moon was the Virtual Rocket Band.''}
    \item \textbf{Hard Distractor:} \textit{``The first band to play on Venus was the Virtual Rocket Band.''}
\end{itemize}
When queried with \textit{``Which band was the first to perform on the Moon?''}, a model relying purely on dense keyword overlap will struggle to distinguish between the two sentences, thereby rigorously testing its true relational and entity-level comprehension.

\paragraph{Visual Needles and Distractors}
To evaluate visual discrimination without the risk of pre-training data leakage, we utilize generative models to synthesize a series of unique, out-of-distribution images (i.e., distinctive objects or scenes that do not exist in reality). These synthetic images serve as our visual needles and are strategically embedded within a vast haystack of randomly sampled background images. To mirror the textual interference strategy, the background context is also populated with visually analogous distractor images. This setup compels the embedding models to rely strictly on zero-shot, fine-grained visual alignment rather than global, coarse-grained scene recognition.

\subsection{Multi-Source Reasoning}
This section details the technical implementation of the construction pipeline for the Multi-Source Reasoning dataset. The complete system prompt utilized during the query filtration stage is provided in Table~\ref{tab:prompt_data_auditor}.

We utilize an upgraded coherent document padding mechanism. Instead of concatenating fragmented sentences to expand the context to standard lengths (8K, 16K, and 32K tokens), we pad with full, contextually intact Wikipedia articles or multimodal document blocks (for the MMDocRAG paradigm). This approach preserves semantic fluency and prevents models from identifying distractors based on unnatural syntactic breaks.

We achieve precise token-level length control using the Qwen3 tokenizer. Crucially, when key evidence sentences are removed to create hard negative samples (e.g., deleting the first, last, or intermediate reasoning hops), they are replaced with coherent text or multimodal segments of strictly equivalent token length. This substitution protocol eliminates potential length bias, ensuring that the embedding models cannot rely on document length as a relevance heuristic. Specific examples of these logic-breaking distractors and the resulting data structures are provided in Table~\ref{tab:appendix_re_case} (here we present only the HotpotQA exemplar, as MMDocRAG follows an identical logic).
\subsection{Logical Summarization}
\label{app:log_summar_details}

This section provides the technical implementation details for the Logical Summarization data generation pipeline, focusing on the specific mechanics used to construct the benchmark.

\paragraph{Mechanics of Global Structure Disruption}
While the main text outlines the strategy of replacing contiguous blocks of the document to destroy global consistency, the source of these replacement blocks (the ``donor'' content) is meticulously controlled to prevent superficial distribution shifts:
\begin{itemize}
    \item \textbf{Text Modality (GovReport):} We utilize a global BM25 index~\cite{Robertson2009ThePR} to retrieve highly similar, yet factually distinct, documents based on the original summary query. When a specific portion of the target text is deleted, we dynamically extract a continuous block of paragraphs from the BM25-retrieved donor document. This ensures the injected noise maintains a deceptively high lexical overlap and domain relevance with the original query, forcing the model to rely strictly on logical flow rather than keyword density.
    \item \textbf{Multimodal Modality (MMLongBench):} For interleaved image--text documents, substituting random external images would create obvious visual discontinuities. Instead, we perform intra-document replacement. We leverage extended versions of the same source document to harvest unseen, continuous image pages. These visually coherent but narratively disjointed pages are injected into the corrupted segments, ensuring that the hard distractors maintain uniform visual formatting and styling throughout.
\end{itemize}

A concrete exemplar for the GovReport dataset is provided in Table~\ref{tab:appendix_log_sum}. The MMLongBench dataset follows the same structural logic and can be referenced similarly; it differs only in modality, as it comprises continuous document images.

\subsection{Temporal Summarization}

We construct a comprehensive benchmark based on ActivityNet Captions and YoukuDenseCaption to rigorously evaluate long-video understanding and temporal reasoning across varying context lengths and retrieval difficulties.

\textbf{Video Sequence Processing.} We extract frames at 1 FPS and stratify both datasets into three context-length buckets (8K, 16K, and 32K), with dataset-specific pool sizes reported in Table~\ref{tab:dataset_stats}. To guarantee clear temporal boundaries when combining contiguous action segments, we apply a strict anti-overlap filter with a 0.5-second margin.

\textbf{Negative Candidate Generation.} Alongside the chronological Positive Candidate, we generate two tiers of negative samples to isolate temporal reasoning from general semantic matching:
\begin{itemize}
    \item \textbf{Hard Negatives (Intra-Video Shuffling):} Visually identical but temporally perturbed sequences (Complete Reversal, Front-Half Shuffle, and Back-Half Shuffle) that force the model to rely strictly on event order.
    \item \textbf{Easy Negatives (Inter-Video Sampling):} Three randomly sampled, distinct videos of matching context length, filtered via a strict blacklist to prevent false negatives.
\end{itemize}

\textbf{Query Temporal Enhancement and Adaptation.} To create strictly temporally binding queries, we prompt Qwen2.5-32B-Instruct to rewrite concatenated captions into cohesive sequences with explicit transition words, while explicitly instructing the model not to introduce unsupported content (Table~\ref{tab:prompt_caption_editor}). Furthermore, to adapt the Chinese annotations in YoukuDenseCaption, we employ the same LLM with greedy decoding as a zero-shot translator. This produces a literal, reproducible English translation while aiming to preserve the original semantics (Table~\ref{tab:prompt_translator}), ensuring the evaluation reflects genuine visual-semantic alignment rather than textual artifacts.

\begin{figure*}[t]
    \centering
    \includegraphics[width=\textwidth]{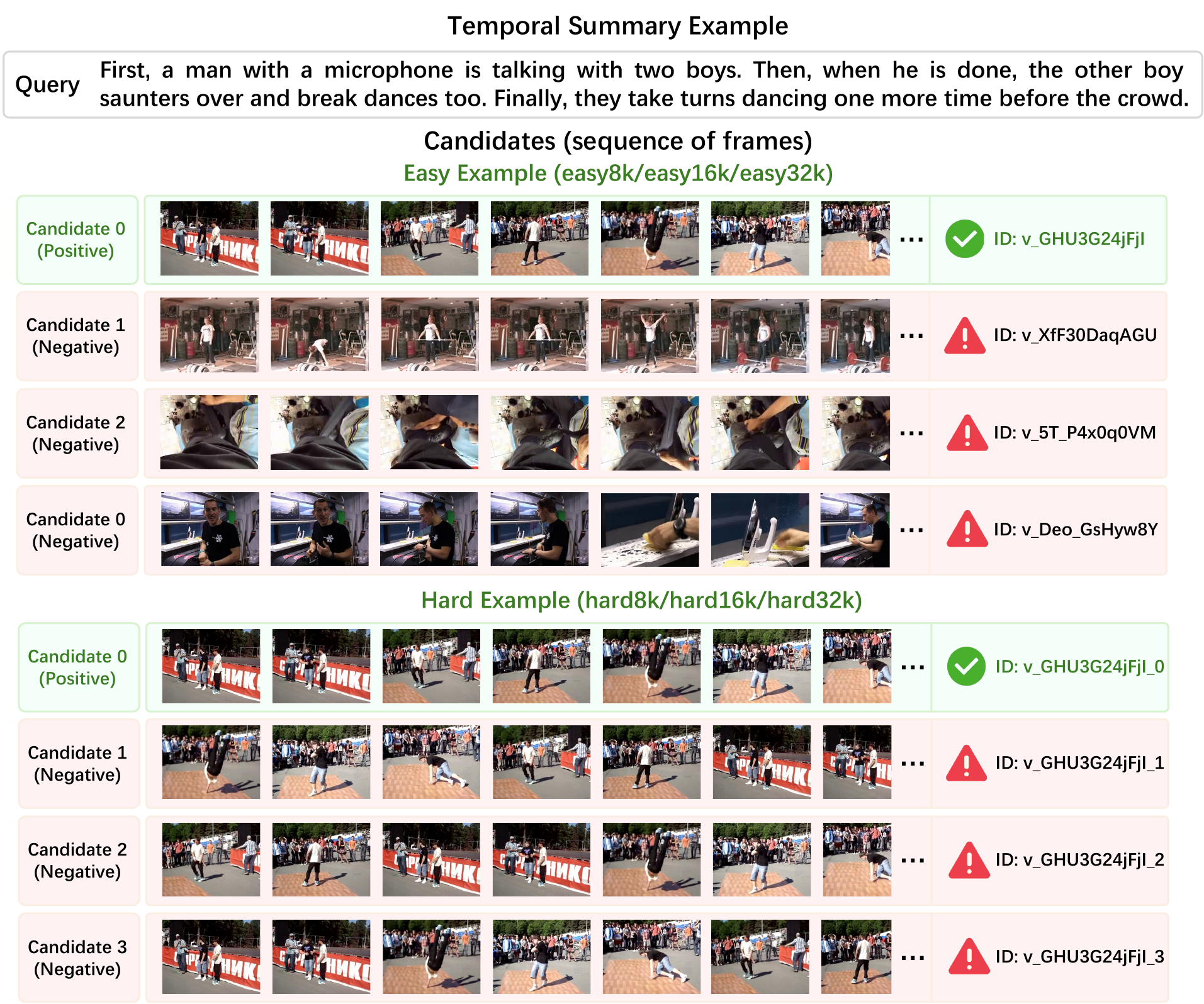}
    \caption{Task formulation for Temporal Summarization. Given a textual query describing a sequence of events, the model is required to retrieve the most relevant video candidate, where each candidate is represented as a sequence of frames. We construct tasks across two datasets, ActivityNet and YoukuDC, under varying context lengths (8K, 16K, 32K) and difficulty levels (easy vs. hard). In the easy setting, negative candidates are sampled from different videos, while in the hard setting, negatives are constructed from the same video with partial temporal overlap and shuffled ordering, requiring fine-grained temporal reasoning. Each task contains one positive (ground-truth) candidate and multiple negatives.}
    \label{fig:tem_sum}
\end{figure*}

We formulate Temporal Summarization as a retrieval task. Given a natural-language query describing a sequence of events, the model ranks a set of candidate videos and retrieves the most relevant one. Each candidate is represented as an ordered sequence of video frames, simulating long-context visual inputs. We construct this benchmark on two datasets, ActivityNet and YoukuDC, and organize the evaluation into six tasks per dataset by varying both input length and difficulty. Specifically, we consider three context lengths (8K, 16K, and 32K tokens) and two difficulty settings (easy and hard). Formally, each instance consists of one query, a set of candidate videos, and a ground-truth label, as shown in Figure~\ref{fig:tem_sum}.

\newcommand{\cmark}{\ding{51}}%
\newcommand{\xmark}{\ding{55}}%

\begin{table*}[t]
\centering
\small
\setlength{\tabcolsep}{5pt}
\begin{tabular}{@{}lccccccc@{}}
\toprule
& \multicolumn{4}{c}{\textbf{Data Source}} & \multicolumn{3}{c}{\textbf{Evaluation Design}} \\
\cmidrule(lr){2-5} \cmidrule(lr){6-8}
\textbf{Benchmark} & \textbf{WebPage} & \textbf{Document} & \textbf{Video} & \textbf{Text} & \textbf{Long Context} & \textbf{Length-Stratified} & \textbf{Hard Distractors} \\
\midrule
LongEmbed & \xmark & \xmark & \xmark & \cmark & \cmark & \xmark & \xmark \\
MMEB      & \cmark & \cmark & \xmark & \xmark & \xmark & \xmark & \xmark \\
MMEB-V2      & \cmark & \cmark & \cmark & \xmark & \xmark & \xmark & \xmark \\
MMTEB     & \xmark & \xmark & \xmark & \cmark & \cmark & \xmark & \xmark \\
LMEB      & \xmark & \xmark & \xmark & \cmark & \xmark & \xmark & \xmark \\
\midrule
\textbf{MMLongEmbed} & \cmark & \cmark & \cmark & \cmark & \cmark & \cmark & \cmark \\
\bottomrule
\end{tabular}
\caption{Comparison of MMLongEmbed with existing embedding benchmarks
across two dimensions: \emph{data source} coverage and \emph{evaluation
design}. Data source categories follow the corpus typology in
Table~\ref{tab:dataset_stats}: \emph{WebPage} denotes interleaved
image--text web content, \emph{Document} paginated visual documents,
\emph{Video} temporal frame sequences, and \emph{Text} text-only corpora.
\emph{Long Context} indicates substantial long-input evaluation;
\emph{Length-Stratified} indicates that context length is systematically
controlled as an independent evaluation variable; \emph{Hard Distractors}
indicates deliberately constructed challenging negatives designed to
suppress superficial matching.}
\label{tab:benchmark_comparison}
\end{table*}
\begin{table*}[t]
\centering
\small
\setlength{\tabcolsep}{5pt}
\resizebox{\textwidth}{!}{%
\begin{tabular}{@{}lccccccc@{}}
\toprule
\textbf{Model} &
\textbf{P@1} &
\textbf{R@5} &
\textbf{R@10} &
\textbf{MRR@5} &
\textbf{MRR@10} &
\textbf{nDCG@5} &
\textbf{nDCG@10} \\
\midrule
Rzen-v2-7B & 58.61 & 82.39 & 84.53 & 68.46 & 68.76 & 71.99 & 72.70 \\
Ops-MM-v1-7B    & 53.90 & 78.44 & 81.15 & 63.96 & 64.33 & 67.62 & 68.51 \\
Qwen3-VL-8B     & 53.41 & 79.26 & 82.01 & 63.90 & 64.27 & 67.78 & 68.68 \\
Qwen3-VL-2B     & 52.31 & 79.26 & 81.85 & 63.15 & 63.51 & 67.21 & 68.06 \\
Doubao-Seed-1.6 & 52.11 & 81.22 & 84.55 & 63.93 & 64.39 & 68.30 & 69.39 \\
Ops-MM-v1-2B    & 48.18 & 75.22 & 78.83 & 58.95 & 59.43 & 63.05 & 64.22 \\
Embed-RL-4B     & 40.84 & 81.19 & 86.71 & 56.46 & 57.21 & 62.68 & 64.47 \\
Embed-RL-2B     & 39.28 & 79.91 & 87.16 & 55.12 & 56.11 & 61.36 & 63.72 \\
VLM2Vec-Qwen2VL-2B         & 31.04 & 54.92 & 58.96 & 40.40 & 40.95 & 44.05 & 45.36 \\
GME-Qwen2-VL-7B & 24.63 & 51.25 & 58.45 & 34.75 & 35.72 & 38.89 & 41.21 \\
GME-Qwen2-VL-2B & 23.10 & 49.70 & 56.00 & 33.42 & 34.26 & 37.50 & 39.56 \\
\bottomrule
\end{tabular}%
}
\caption{
Supplementary retrieval metrics on MMLongEmbed.
P@1 denotes Precision@1; R@$k$, MRR@$k$, and nDCG@$k$ denote
Recall, Mean Reciprocal Rank, and normalized Discounted Cumulative Gain
at cutoff $k$, respectively.
}
\label{tab:full_metrics}
\end{table*}
\begin{table*}[t]
    \centering
    \resizebox{\textwidth}{!}{
    \begin{tabular}{lll}
        \toprule
        \textbf{Model Name} & \textbf{Model Version} & \textbf{Model Link} \\
        \midrule
        \multicolumn{3}{l}{\textit{Open-Source / Local Models}} \\
        Qwen3-VL-2B & Qwen3-VL-Embedding-2B & https://huggingface.co/Qwen/Qwen3-VL-Embedding-2B \\
        Qwen3-VL-8B & Qwen3-VL-Embedding-8B & https://huggingface.co/Qwen/Qwen3-VL-Embedding-8B \\
        GME-Qwen2-VL-2B & GME-Qwen2-VL-2B-Instruct & https://huggingface.co/Alibaba-NLP/gme-Qwen2-VL-2B-Instruct \\
        GME-Qwen2-VL-7B & GME-Qwen2-VL-7B-Instruct & https://huggingface.co/Alibaba-NLP/gme-Qwen2-VL-7B-Instruct \\
        Rzen-v2-7B & RzenEmbed-v2-7B & https://huggingface.co/qihoo360/RzenEmbed \\
        Ops-MM-v1-2B & Ops-MM-embedding-v1-2B & https://huggingface.co/OpenSearch-AI/Ops-MM-embedding-v1-2B \\
        Ops-MM-v1-7B & Ops-MM-embedding-v1-7B & https://huggingface.co/OpenSearch-AI/Ops-MM-embedding-v1-7B \\
        Embed-RL-2B & Embed-RL-2B & https://huggingface.co/ZoengHouNaam/Embed-RL-2B \\
        Embed-RL-4B & Embed-RL-4B & https://huggingface.co/ZoengHouNaam/Embed-RL-4B \\
        VLM2Vec-Qwen2VL-2B & VLM2Vec-Qwen2VL-2B & https://huggingface.co/TIGER-Lab/VLM2Vec-Qwen2VL-2B \\
        \midrule
        \multicolumn{3}{l}{\textit{Proprietary Models}} \\
        Doubao-Seed-1.6 & seed1.6-embedding-1215 & - \\
        \bottomrule
    \end{tabular}}
    \caption{Detailed information of models evaluated in MMLongEmbed.}
    \label{tab:model_info}
\end{table*}
\begin{table*}[t]
\centering
\small
\setlength{\tabcolsep}{5pt}
\begin{tabular}{@{}llrrrrrrr@{}}
\toprule
\textbf{Family} &
\textbf{Large vs.\ Small} &
\textbf{Large P@1} &
\textbf{Small P@1} &
\textbf{Diff.} &
\textbf{95\% CI} &
\textbf{$t$} &
\textbf{$p$} &
\textbf{Cohen's $d$} \\
\midrule
Qwen3
& 8B vs.\ 2B
& 53.98 & 53.65 & +0.33
& [-0.72, 1.36] & 0.62 & 0.5375 & 0.007 \\

Ops-MM
& 7B vs.\ 2B
& 59.04 & 53.32 & +5.72
& [4.65, 6.80] & 10.42 & $<0.0001$ & 0.113 \\

Embed-RL
& 4B vs.\ 2B
& 41.83 & 39.85 & +1.98
& [1.06, 2.92] & 4.15 & $<0.0001$ & 0.045 \\

GME
& 7B vs.\ 2B
& 25.89 & 24.15 & +1.74
& [0.89, 2.59] & 4.02 & 0.0001 & 0.044 \\
\bottomrule
\end{tabular}

\caption{
Statistical comparison of model-scale effects within each model family.
Paired $t$-tests are conducted on query-level P@1 outcomes between the
largest and smallest variants. Diff.\ denotes Large minus Small.
The 95\% confidence intervals are estimated using 10{,}000 bootstrap
resamples of paired differences, and Cohen's $d$ reports the standardized
effect size. P@1 values here are query-level micro-averages, whereas the Avg.\
scores in Table~\ref{tab:model_benchmark} are macro-averages over the 14 dataset
subsets with equal subset weighting; thus, the values may differ slightly
between the two tables.
}
\label{tab:scale_tests}
\end{table*}
\begin{table*}[t]
\centering
\small
\setlength{\tabcolsep}{4pt}
\resizebox{\textwidth}{!}{%
\begin{tabular}{@{}lccccc@{}}
\toprule
\textbf{Model} &
\textbf{Depth Means (\%)} &
\textbf{LMM Wald $\chi^2(5)$} &
\textbf{$p$} &
\textbf{Quadratic $\chi^2(1)$} &
\textbf{Quadratic $p$} \\
\midrule

Rzen-v2-7B
& 95.9, 80.0, 78.7, 77.2, 77.4, 80.9
& 101.08 & $<0.0001$ & 50.38 & $<0.0001$ \\

Qwen3-VL-8B
& 84.8, 72.6, 68.5, 67.4, 68.3, 75.0
& 63.80 & $<0.0001$ & 45.91 & $<0.0001$ \\

Ops-MM-v1-7B
& 93.3, 77.8, 77.4, 78.3, 78.3, 82.6
& 73.53 & $<0.0001$ & 46.00 & $<0.0001$ \\

Doubao-Seed-1.6
& 93.9, 72.6, 70.2, 74.4, 70.7, 83.5
& 147.75 & $<0.0001$ & 108.90 & $<0.0001$ \\

Qwen3-VL-2B
& 92.0, 70.4, 67.4, 66.9, 65.7, 72.2
& 146.64 & $<0.0001$ & 77.90 & $<0.0001$ \\

Ops-MM-v1-2B
& 77.4, 72.4, 73.3, 71.9, 73.3, 88.3
& 61.93 & $<0.0001$ & 38.73 & $<0.0001$ \\

Embed-RL-4B
& 96.7, 59.4, 58.3, 59.1, 57.6, 57.6
& 339.95 & $<0.0001$ & 110.50 & $<0.0001$ \\

Embed-RL-2B
& 90.6, 58.7, 54.4, 56.1, 54.4, 53.7
& 275.01 & $<0.0001$ & 85.39 & $<0.0001$ \\

VLM2Vec-Qwen2VL-2B
& 56.7, 59.6, 62.2, 64.4, 64.3, 80.9
& 86.32 & $<0.0001$ & 9.53 & $0.0020$ \\

GME-Qwen2-VL-7B
& 73.7, 53.1, 51.5, 52.2, 50.9, 53.0
& 91.62 & $<0.0001$ & 36.39 & $<0.0001$ \\

GME-Qwen2-VL-2B
& 69.3, 52.4, 50.4, 49.3, 50.6, 49.1
& 70.60 & $<0.0001$ & 22.30 & $<0.0001$ \\

\bottomrule
\end{tabular}%
}

\caption{
Statistical analysis of positional bias on Visual Grounding.
Linear mixed-effects models (LMMs) evaluate the effect of six insertion
depths on P@5, with needle instance as a random intercept.
Depth means are reported from the beginning to the end of the sequence.
LMM Wald $\chi^2(5)$ tests the overall depth effect, while Quadratic
$\chi^2(1)$ and its $p$-value test the quadratic component of the depth
trend. The 0.25K and 0.5K length bins are excluded.
}
\label{tab:position_tests}
\end{table*}

\begin{table*}[htbp]
\centering
\small
\setlength{\tabcolsep}{0pt}
\renewcommand{\arraystretch}{1.2}
\begin{tabular}{@{}p{\linewidth}@{}}
\toprule
\multicolumn{1}{@{}l@{}}{\textbf{Logical Summarization GovReport Case}} \\
\midrule
\textbf{Query:} Given this summary, retrieve the corresponding original long document. \\
(Summary covers: GAO report on military burn pits in Afghanistan/Iraq, DOD compliance, alternatives, and health monitoring.) \\[0.4em]

\textbf{Candidate 0 (Ground Truth)} \\
GAO was asked to report on the (1) extent of open pit burning in Afghanistan and Iraq, and whether the military has followed its guidance; (2) alternatives to burn pits, and whether the military has examined them; and (3) extent of efforts to monitor air quality and potential health impacts. CENTCOM did not develop comprehensive burn pit guidance until 2009. Operators of burn pits at four bases GAO visited in Iraq were not complying with key elements of this guidance. DOD has been slow to implement alternatives or fully evaluate their benefits and costs. U.S. Forces do not sample or monitor burn pit emissions as provided by key CENTCOM regulation. \\
Coverage: \textbf{Complete.} Addresses all three core dimensions: regulatory compliance, waste management alternatives, and health impact monitoring. \\[0.5em]

\textbf{Distractor 1 (Partial $\cdot$ Missing [Alternatives] \& [Health Monitoring])} \\
Between January and March 2010, we determined that, to varying degrees, the four burn pits we visited at bases in Iraq were not managed in accordance with CENTCOM's 2009 regulation. For example, operators at all four of these burn pits burned varying amounts of plastic—a prohibited item that can produce carcinogens when burned. At Warhorse, despite some limited waste sorting efforts, a burn pit operator said they did not segregate plastic from the waste stream. \\
Coverage: \textbf{Partial.} Covers compliance failures only. \textbf{Missing:} systematic evaluation of disposal alternatives and long-term health monitoring efforts. \\[0.5em]

\textbf{Distractor 2 (Partial $\cdot$ Missing [Compliance] \& [Health Monitoring])} \\
The U.S. military relies on civilian contractors to provide supplies and services, including managing some burn pits, in support of its contingency operations in Afghanistan and Iraq. Kellogg, Brown, and Root (KBR) has provided burn pit services in Iraq through the Logistics Civil Augmentation Program (LOGCAP) III contract. Typically, contractors such as KBR, DynCorp, and Fluor work under task orders. \\
Coverage: \textbf{Partial.} Details contractor logistics frameworks. \textbf{Missing:} guidance adherence audit and health impact assessment required by the summary. \\[0.5em]

\textbf{Distractor 3 (Partial $\cdot$ Missing [Compliance] \& [Alternatives])} \\
Air pollution in Afghanistan and Iraq is generally high. For example, the level of particulate matter is higher in Afghanistan and Iraq than in the United States. Particulate matter includes coarse particles between 2.5 and 10 micrometers in diameter, as well as fine particles smaller than 2.5 micrometers. Health problems associated with particle pollution identified by EPA include irritation of the airways, coughing, or difficulty breathing; decreased lung function; aggravated asthma. \\
Coverage: \textbf{Partial.} Covers general air quality background. \textbf{Missing:} specific regulatory compliance audit and alternative disposal analysis. \\[0.4em]

\bottomrule
\end{tabular}
\caption{Logical 
case design. Unlike multi-hop reasoning tasks, this evaluation focuses on summary coverage completeness. The ground truth addresses the full audit scope, while distractors exhibit fragmented coverage by isolating specific sub-themes. This design tests embedding models' ability to assess macro-level content alignment rather than relying on localized keyword overlap.}
\label{tab:appendix_log_sum}
\end{table*}
\begin{table*}[htbp]
\centering
\small
\setlength{\tabcolsep}{0pt}
\renewcommand{\arraystretch}{1.25}
\begin{tabular}{@{}p{\linewidth}@{}}
\toprule
\multicolumn{1}{@{}l@{}}{\textbf{Multi-Source Reasoning HotpotQA Case}} \\
\midrule
\textbf{Query:} What was the third album Nathan Chapman produced for Taylor Swift, released in late 2010? \\[0.5em]

\textbf{Candidate 0 (Ground Truth)} \\
``Sparks Fly'' is a song written and recorded by American singer-songwriter Taylor Swift for her third studio album \textbf{``Speak Now'' (2010)}. The song was written by Swift and Liz Rose and produced by \textbf{Nathan Chapman}, with Swift's aid. It was released on \textbf{October 25, 2010}, by Big Machine Records. \\
\textit{Chain Status: Intact. Preserves all reasoning nodes: [Producer] $\rightarrow$ [Ordinal] $\rightarrow$ [Time] $\rightarrow$ [Entity].} \\[0.6em]

\textbf{Distractor 1 (Hard $\cdot$ Temporal Link Removed)} \\
``Sparks Fly'' is a song written and recorded by American singer-songwriter Taylor Swift for her third studio album ``Speak Now'' (2010). The song was written by Swift and Liz Rose and produced by Nathan Chapman, with Swift's aid. \textbf{The album was released on November 11, 2008}, by Big Machine Records. Swift was 16 years old at the time of the album's release and wrote its songs during her freshman year of high school. \\
\textit{Chain Status: Broken. \textbf{Removes the [Time] reasoning frame.} Replaces ``late 2010'' with 2008, creating a temporal-ordinal contradiction that fractures the chronological verification chain.} \\[0.6em]

\textbf{Distractor 2 (Easy $\cdot$ Entity/Domain Link Removed)} \\
The ABC Mosquito was a 120 hp (90 kW) six-cylinder radial aero engine designed by British engineer Granville Bradshaw for light aircraft. The single prototype was built by ABC Motors and first ran in 1916. Captain Sir Geoffrey de Havilland (1882--1965) was a British aviation pioneer. His \textbf{Mosquito} has been considered the most versatile warplane of World War II. \\
\textit{Chain Status: Broken. \textbf{Removes the [Entity/Domain] reasoning frame.} Deletes all music-industry entities (Taylor Swift, Nathan Chapman) and shifts context to aviation, severing the core subject chain.} \\[0.6em]

\bottomrule
\end{tabular}
\caption{Multi-source reasoning case design for HotpotQA. The hard distractor preserves topical overlap while breaking a key temporal link, whereas the easy distractor shifts to an unrelated entity/domain. This contrast illustrates how the two difficulty settings distinguish chain-level inconsistency from unrelated negatives.}
\label{tab:appendix_re_case}
\end{table*}
\begin{table*}[htbp]
\centering
\small
\setlength{\tabcolsep}{6pt}
\begin{tabular}{@{}p{\linewidth}@{}}
\toprule
\multicolumn{1}{@{}l@{}}{\textbf{System Prompt: Senior Data Auditor for Multi-Source Reasoning Data}} \\
\midrule
\textbf{Role \& Task:} Evaluate if the input \texttt{QUERY} contains a ``Strongly Linked Anchor'' capable of uniquely identifying EXACTLY ONE target document within a large-scale corpus. \\[0.4em]

\textbf{Evaluation Logic:}
\begin{enumerate}
    \item \textbf{Extract Candidates.} Identify high-specificity signals:
    \begin{itemize}
        \item Specific Entities: Proper nouns, unique IDs, model names (e.g., ``Llama-3-70B'', ``Project Zephyr-X'').
        \item Unique Events: Time-bound incidents, named operations, rare occurrences (e.g., ``2023 Maui wildfire evacuation protocol'', ``Operation Nightfall data leak'').
        \item Distinctive Data: Non-generic statistics, experimental results, or metrics unlikely to repeat (e.g., ``accuracy=98.73\% on MMLU-Pro'', ``patient cohort N=1,247 with BRCA1-R1699Q'').
        \item Rare Combinations: Uncommon conjunctions of time + location + subject + metric (e.g., ``Q3 2024 Shanghai EV battery recycling rate'').
    \end{itemize}
    \item \textbf{Filter Broad Signals.} Reject if candidate is:
    \begin{itemize}
        \item Too generic (``AI'', ``the study'', ``2024'', ``user feedback'').
        \item Recurring pattern without unique modifiers (``a clinical trial'', ``an earthquake response'').
        \item Ambiguous without external context.
    \end{itemize}
    \item \textbf{Uniqueness Validation.} Ask: ``If I search this anchor in a 10M-doc corpus, would $\ge 95\%$ of retrieved docs be the SAME target?'' $\rightarrow$ Only ``Yes'' qualifies as valid.
\end{enumerate}

\textbf{Output Specification (Strict JSON):} \\
\begin{minipage}{\linewidth}
\begin{verbatim}
{
  "is_valid": true/false,
  "strong_linked_anchor": "The exact phrase identified (noun/event/data/combination)",
  "anchor_type": "entity/event/data/combination",
  "potential_ambiguity": "High/Low",
  "reason": "One concise sentence: why this anchor uniquely identifies one doc"
}
\end{verbatim}
\end{minipage}
\end{tabular}
\caption{System prompt for the Senior Data Auditor, designed to systematically evaluate the presence and discriminative power of Strongly Linked Anchors in retrieval queries.}
\label{tab:prompt_data_auditor}
\end{table*}
\begin{table*}[htbp]
\centering
\small
\setlength{\tabcolsep}{6pt}
\begin{tabular}{@{}p{\linewidth}@{}}
\toprule
\multicolumn{1}{@{}l@{}}{\textbf{System Prompt: Expert Video Caption Editor for Temporal Sequencing}} \\
\midrule
\textbf{Role \& Task:} You are an expert video caption editor focused on clarifying action sequences. Your task is to rewrite a given \texttt{original\_caption} to explicitly emphasize the chronological order of events. \\[0.4em]

\textbf{Rewriting Guidelines:}
\begin{enumerate}
    \item \textbf{Preserve Meaning.} Maintain the exact original meaning. Do NOT add any new events, objects, or details that are not present in the original text.
    \item \textbf{Enhance Temporal Sequence.} Structure the caption to clearly highlight what happens first and what happens later. Use explicit transition words, such as:
    \begin{itemize}
        \item First, Initially
        \item Then, Next, Afterwards
        \item Finally
    \end{itemize}
\end{enumerate}

\textbf{Output Specification (Strict Text):} \\
\begin{itemize}
    \item Output \textbf{ONLY} the rewritten caption text.
    \item Do \textbf{NOT} include any introductory phrases, explanations, or quotes.
\end{itemize}

\textbf{Input Format:} \\
\begin{minipage}{\linewidth}
\begin{verbatim}
Original caption: {original_caption}
\end{verbatim}
\end{minipage} \\
\bottomrule
\end{tabular}
\caption{System prompt for the Expert Video Caption Editor, designed to emphasize the chronological order of events while constraining the model from introducing unsupported content.}
\label{tab:prompt_caption_editor}
\end{table*}
\begin{table*}[htbp]
\centering
\small
\setlength{\tabcolsep}{6pt}
\begin{tabular}{@{}p{\linewidth}@{}}
\toprule
\multicolumn{1}{@{}l@{}}{\textbf{System Prompt: Professional Translator for Exact Translation}} \\
\midrule
\textbf{Role \& Task:} You are a professional translator. Your task is to translate the provided Chinese \texttt{text} into English while maintaining strict fidelity to the original source. \\[0.4em]

\textbf{Translation Rules:}
\begin{itemize}
    \item \textbf{Strict Fidelity:} Keep ALL meaning exactly unchanged.
    \item \textbf{No Additions:} Do NOT add any new information or context.
    \item \textbf{No Omissions:} Do NOT remove any information from the original text.
    \item \textbf{No Paraphrasing:} Do NOT paraphrase; maintain a direct and exact translation.
\end{itemize}

\textbf{Output Specification (Strict Text):} \\
\begin{itemize}
    \item Output \textbf{ONLY} the translation.
    \item Do not include any introductory phrases, explanations, or conversational filler.
\end{itemize}

\textbf{Input Format:} \\
\begin{minipage}{\linewidth}
\begin{verbatim}
Text:
{text}
\end{verbatim}
\end{minipage} \\
\bottomrule
\end{tabular}
\caption{System prompt for the Professional Translator, designed to preserve the source meaning as faithfully as possible by prohibiting additions, omissions, and free paraphrasing.}
\label{tab:prompt_translator}
\end{table*}

\section{Detailed Evaluation Results}
\label{detailed_evaluation_results}
Detailed specifications of the evaluated models are summarized in Table~\ref{tab:model_info}. In this section, we provide comprehensive empirical results and extended visualizations to complement the main analysis. To ensure transparency and reproducibility, we report both quantitative performance metrics and qualitative visualizations across all evaluated multimodal models.

\subsection{Extended Visualizations for Context Length and Insertion Depth}
\label{subsec:extended_visualizations}

To offer an exhaustive overview of how varying context lengths and target insertion depths impact empirical retrieval accuracy, we aggregate the fine-grained visual profiles for all candidate models.

\begin{figure*}[htbp]
    \centering
    \includegraphics[width=\textwidth]{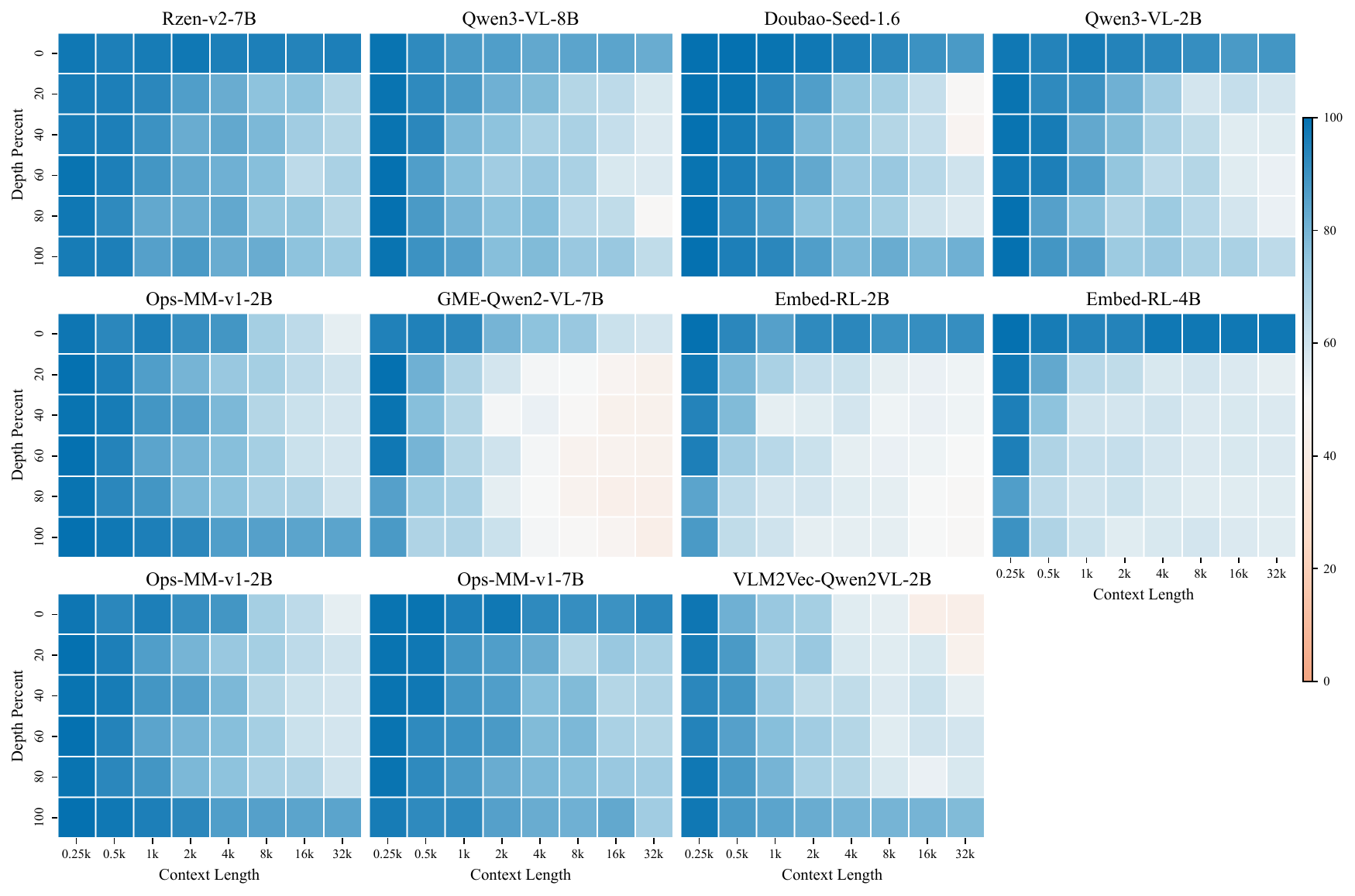}
    \caption{Complete visualization of model performance w.r.t. insertion depth and context length in the visual grounding task across all evaluated models.}
    \label{fig:heatmap_vg_full}
\end{figure*}

Specifically, Figure~\ref{fig:heatmap_vg_full} illustrates the complete visual evaluation profiles, depicting each model's performance boundaries with respect to insertion depth and context length in the visual grounding task. This extended visualization serves as a macro-level supplement to Figure~\ref{fig:heatmap_vg} in the main text, capturing localized performance degradations across all tested multimodal configurations.

\subsection{Quantitative Modality Bias Metrics}
\label{subsec:quantitative_metrics}
While the visual heatmaps illustrate macro-level behavioral trends, this subsection provides the corresponding numerical results for quantitative comparison. Table~\ref{tab:modality_bias} reports performance under text-only, mixed-modal, and image-only background interference, grouped by modality interaction pattern. Mean ($\mu$) and sample standard deviation ($\sigma$) summarize overall performance and cross-setting variation, respectively. These metrics are defined as:
\begin{equation}
    \mu = \frac{1}{N}\sum_{i=1}^{N}x_i
\end{equation}
\begin{equation}
    \sigma = \sqrt{\frac{1}{N-1}\sum_{i=1}^{N}(x_i - \mu)^2}
\end{equation}
where $N$ represents the total number of evaluated background conditions and $x_i$ denotes the corresponding performance score. This complete reference retains the prominent models from the main text as empirical anchors while incorporating additional baseline configurations, providing a solid statistical foundation for the modality bias observed in the visual grounding task.

\begin{table*}[htbp]
\centering
\small
\renewcommand{\arraystretch}{1.2}
\setlength{\tabcolsep}{2pt}

\resizebox{\linewidth}{!}{%
\begin{tabular}{
>{\raggedright\arraybackslash}p{0.38\linewidth}
>{\centering\arraybackslash}p{0.11\linewidth}
>{\centering\arraybackslash}p{0.11\linewidth}
>{\centering\arraybackslash}p{0.11\linewidth}
>{\centering\arraybackslash}p{0.14\linewidth}
>{\centering\arraybackslash}p{0.15\linewidth}
}
\toprule
\textbf{Model} & \textbf{Text} & \textbf{Mix} & \textbf{Image} & \textbf{Mean ($\mu$)} & \textbf{Std. Dev. ($\sigma$)} \\
\midrule

\rowcolor{gray!15}
\multicolumn{6}{@{}l@{}}{\textit{Category 1: Modality-Skewed Models}} \\
Qwen3-VL-2B         & 63.19 & 82.85 & 88.61 & 78.22 & 13.33 \\
Qwen3-VL-8B         & 64.17 & 81.67 & 89.03 & 78.29 & 12.77 \\
Ops-MM-v1-2B        & 75.28 & 80.14 & 88.91 & 81.44 & 6.91  \\
GME-Qwen2-VL-7B     & 55.69 & 62.08 & 72.99 & 63.59 & 8.75  \\
VLM2Vec-Qwen2VL-2B  & 58.19 & 70.28 & 85.51 & 71.33 & 13.69 \\
Doubao-Seed-1.6     & 75.83 & 87.15 & 85.14 & 82.71 & 6.03  \\

\midrule

\rowcolor{gray!15}
\multicolumn{6}{@{}l@{}}{\textit{Category 2: Cross-Modal Interference Sensitive Models}} \\

Embed-RL-2B         & 65.49 & 61.94 & 73.61 & 67.01 & 6.00  \\
Embed-RL-4B         & 71.81 & 62.43 & 75.35 & 69.86 & 6.68  \\

\midrule

\rowcolor{gray!15}
\multicolumn{6}{@{}l@{}}{\textit{Category 3: Modality-Robust Models}} \\
Rzen-v2-7B          & 86.94 & 82.71 & 86.32 & 85.32 & 2.29  \\
Ops-MM-v1-7B        & 82.50 & 83.54 & 89.51 & 85.18 & 3.83  \\

\bottomrule
\end{tabular}%
}

\caption{Performance of MEMs under heterogeneous background interference in \emph{Visual Grounding}. The table reports Precision@1 under text, mixed, and image distractors. Models are grouped by modality interaction patterns: \textit{modality-skewed} (uneven performance across modalities), \textit{interference-sensitive} (cross-modal inputs degrade performance), and \textit{modality-robust} (stable performance across all settings). Mean ($\mu$) and standard deviation ($\sigma$) quantify overall performance and cross-setting robustness, respectively.
}
\label{tab:modality_bias_full}
\end{table*}

\subsection{Detailed Scores Across Varying Lengths}
While Section~\ref{sec:analysis_realworld} visualizes the performance degradation trajectories across different context scales, the exact numerical values were omitted for visual clarity. To provide a comprehensive quantitative reference, we report the detailed scores of all evaluated models across various tasks and lengths. As shown in Table~\ref{tab:length_details}, this granular breakdown allows for a precise comparison of model capabilities under diverse context windows and difficulty settings.
\begin{table*}[htbp]
\centering
\begin{tabular}{l ccc ccc}
\toprule
\multirow{2}{*}{\textbf{Model}} & \multicolumn{3}{c}{\textbf{EASY}} & \multicolumn{3}{c}{\textbf{HARD}} \\
\cmidrule(lr){2-4} \cmidrule(lr){5-7}
& \textbf{8k} & \textbf{16k} & \textbf{32k} & \textbf{8k} & \textbf{16k} & \textbf{32k} \\
\midrule
\multicolumn{7}{c}{\textit{Multi-Source Reasoning}} \\
\midrule
Rzen-v2-7B         & \textbf{77.19} & \textbf{65.00} & \textbf{58.13} & \textbf{71.88} & \textbf{57.50} & \textbf{50.94} \\
Qwen3-VL-8B        & 70.63 & \underline{59.38} & 52.81 & \underline{64.38} & 45.63 & 38.44 \\
Qwen3-VL-2B        & 68.75 & 59.06 & 42.19 & 57.19 & \underline{51.25} & 36.56 \\
Doubao-Seed-1.6    & 61.25 & 34.38 & 23.44 & 58.13 & 33.75 & 20.63 \\
Ops-MM-v1-7B       & \underline{74.38} & 56.88 & 50.94 & 62.81 & 44.38 & \underline{39.69} \\
Ops-MM-v1-2B       & 71.56 & 47.50 & 39.69 & 54.06 & 32.50 & 25.31 \\
GME-Qwen2-VL-7B    & 20.94 &  5.63 &  4.38 & 15.63 &  3.75 &  3.44 \\
GME-Qwen2-VL-2B    & 20.31 &  6.25 &  5.63 & 16.88 &  4.06 &  2.19 \\
VLM2Vec-Qwen2VL-2B & 23.75 &  1.25 &  0.31 & 22.50 &  2.19 &  0.94 \\
Embed-RL-4B        & 54.38 & 56.25 & \underline{53.44} & 26.56 & 27.81 & 27.19 \\
Embed-RL-2B        & 51.56 & 51.88 & 51.25 & 27.81 & 24.38 & 25.63 \\
\midrule
\multicolumn{7}{c}{\textit{Logical Summarization}} \\
\midrule
Rzen-v2-7B         & \underline{91.88} & \textbf{98.96} & 77.50 & \underline{81.04} & \underline{67.50} & \underline{61.04} \\
Qwen3-VL-8B        & 90.21 & 95.63 & \underline{79.38} & 74.58 & \textbf{67.71} & \textbf{61.67} \\
Qwen3-VL-2B        & 89.17 & 96.25 & \textbf{79.79} & \textbf{86.04} & 62.71 & 56.04 \\
Doubao-Seed-1.6    & 87.92 & \underline{97.08} & 75.83 & 72.71 & 56.67 & 45.00 \\
Ops-MM-v1-7B       & \textbf{95.00} & 95.83 & 60.00 & 60.83 & 45.00 & 42.50 \\
Ops-MM-v1-2B       & 91.46 & 95.21 & 73.33 & 70.21 & 57.50 & 60.21 \\
GME-Qwen2-VL-7B    & 48.75 & 43.54 & 46.46 & 18.33 & 15.42 & 21.46 \\
GME-Qwen2-VL-2B    & 51.25 & 47.50 & 44.58 & 22.71 & 13.96 & 15.63 \\
VLM2Vec-Qwen2VL-2B & 39.38 & 50.42 & 68.96 & 40.42 & 32.92 & 50.63 \\
Embed-RL-4B        & 76.46 & 80.42 & 69.58 & 30.00 & 24.17 & 31.25 \\
Embed-RL-2B        & 72.08 & 77.50 & 63.33 & 32.29 & 28.96 & 29.58 \\
\midrule
\multicolumn{7}{c}{\textit{Temporal Summarization}} \\
\midrule
Rzen-v2-7B         & \textbf{92.72} & 86.17 & 72.31 & \underline{33.94} & \textbf{26.69} & \textbf{42.69} \\
Qwen3-VL-8B        & 88.83 & \underline{87.86} & \underline{92.33} & 23.89 & 25.19 & 30.39 \\
Qwen3-VL-2B        & 84.97 & 81.39 & 91.94 & 22.14 & 25.50 & 18.25 \\
Doubao-Seed-1.6    & \underline{91.50} & \textbf{89.92} & \textbf{95.11} & 21.11 & 23.42 & 31.78 \\
Ops-MM-v1-7B       & 85.75 & 84.67 & 83.39 & \textbf{36.89} & \underline{25.81} & \underline{40.19} \\
Ops-MM-v1-2B       & 77.17 & 70.72 & 80.00 & 21.22 & 16.17 & 27.75 \\
GME-Qwen2-VL-7B    & 53.50 & 37.00 & 48.42 & 10.17 &  9.78 & 10.36 \\
GME-Qwen2-VL-2B    & 43.44 & 36.28 & 35.36 &  4.78 &  7.28 &  5.47 \\
VLM2Vec-Qwen2VL-2B & 73.19 & 72.42 & 79.19 & 22.42 & 13.36 & 28.22 \\
Embed-RL-4B        & 52.61 & 37.58 & 48.72 &  6.28 &  9.17 &  9.67 \\
Embed-RL-2B        & 50.61 & 35.47 & 45.14 &  6.78 &  4.58 &  7.67 \\
\bottomrule
\end{tabular}
\caption{Score variations of different models across tasks under diverse context lengths. Results are stratified into EASY and HARD subsets.}
\label{tab:length_details}
\end{table*}

\end{document}